\documentclass[11pt,letterpaper]{article}
\usepackage{emnlp2017}
\usepackage{times}
\usepackage{latexsym}
\usepackage{amsmath}
\usepackage{amssymb}
\usepackage{mathtools}
\usepackage{multirow}
\usepackage{enumerate}

\newcommand{\neatlist}{\setlength{\parskip}{0ex}\setlength{\itemsep}{1ex}\setlength{\parsep}{0ex}}

\newcommand{\cev}[1]{\reflectbox{\ensuremath{\vec{\reflectbox{\ensuremath{#1}}}}}}
\emnlpfinalcopy

\def\emnlppaperid{***}

\title{
Zero-Shot Activity Recognition with Verb Attribute Induction
}

\author{Rowan Zellers \and Yejin Choi \\
  Paul G. Allen School of Computer Science \& Engineering \\
  University of Washington\\
  Seattle, WA 98195, USA \\
  {\tt \{rowanz,yejin\}@cs.washington.edu}}

\date{}

\begin{document}

\maketitle
\begin{abstract}
In this paper, we investigate large-scale zero-shot activity recognition by modeling the visual and linguistic attributes of action verbs. For example, the verb ``salute'' has several properties, such as being a light movement, a social act, and short in duration.
We use these attributes as the internal mapping between visual and textual representations to reason about a previously unseen action.
In contrast to much prior work that assumes access to gold standard attributes for zero-shot classes and focuses primarily on object attributes, our model uniquely learns to infer action attributes from dictionary definitions and distributed word representations. Experimental results confirm that action attributes inferred from language can provide a predictive signal for zero-shot prediction of previously unseen activities.
%We introduce the task of inferring action verb attributes from how the word is defined and used in context. For example, given a verb such as ``salute'', we want to infer visual and linguistic properties of the action such as
%motion dynamics (light movement), social dynamics (social act), duration (short),
%and Vendler's aspectual classes (achievement).
%We develop a new verb attribute dataset and establish baseline performance in this task using models that can learn to predict attributes from dictionary definitions and distributed word representations.
%To probe empirical use of verb attributes, we investigate zero-shot activity recognition from images, where verb attributes are used as the internal mappings between the visual and textual representations. Experimental results confirm that verb attributes can be inferred from language and that they provide predictive signals for zero-shot prediction of previously unseen actions.
\end{abstract}
%%%%%%%%%%%%%%%%%%%%%%%%%%%%%%%%%%%%%%%%%%%%%%%%%%%%%%%%%%%%%%%%%%%%%
\section{Introduction}
\label{sec:intro}

\begin{figure}[t!]
\centerline{\includegraphics[scale=0.70]{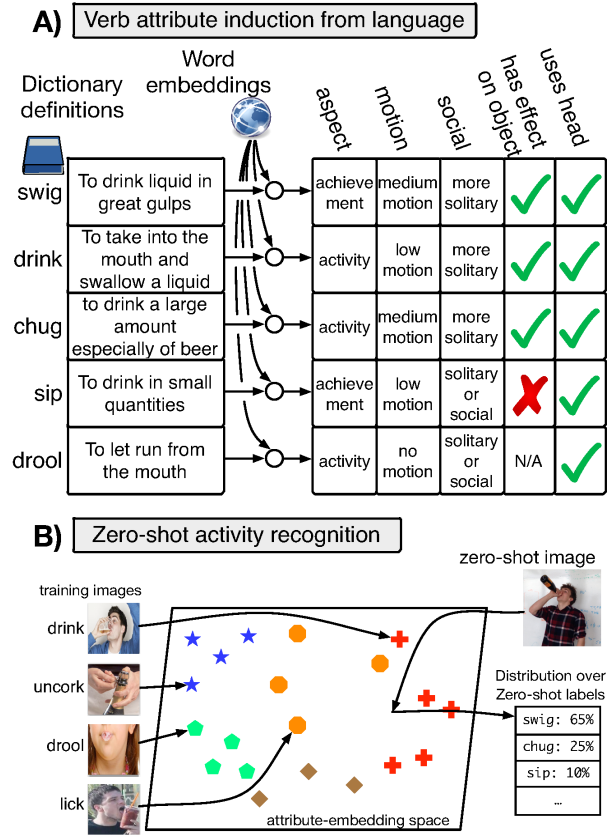}}
\caption{An overview of our task. Our goal is twofold. \textbf{A:} we seek to use use distributed word embeddings in tandem with dictionary definitions to obtain a high level understanding of verbs.
\textbf{B:} we seek to use these predicted attributes to allow a classifier to recognize a broader set of activities than what was seen in training time.}
\label{fig:overview}
\end{figure}

We study the problem of inferring action verb attributes based on how the word is defined and used in context. For example, given a verb such as ``swig'' shown in Figure~\ref{fig:overview}, we want to infer various properties of actions such as \emph{motion dynamics} (moderate movement), \emph{social dynamics} (solitary act), \emph{body parts} involved (face, arms, hands), and \emph{duration} (less than 1 minute) that are generally true for the range of actions that can be denoted by the verb ``swig.''

Our ultimate goal is to improve \emph{zero-shot learning} of activities in computer vision: predicting a previously unseen activity by integrating %non-visual background knowledge about the space of possible activities and objects.
background knowledge about the conceptual properties of actions.
For example, a computer vision system may have seen images of ``drink'' activities during training, but not ``swig.'' Ideally, the system should infer the likely visual characteristics of ``swig'' using world knowledge implicitly available in dictionary definitions and word embeddings.

However, most existing literature on zero-shot learning has focused on object recognition, with only a few notable exceptions (see Related Work in Section \ref{sec:related}). %Indeed, this is critical:
There are two critical reasons:
%\emph{object attributes}, such as color, shape, and texture, are both conceptually straightforward to enumerate as well as robust for current vision systems to recognize.
\emph{object attributes}, such as color, shape, and texture, are  conceptually straightforward to enumerate. In addition, they have distinct visual patterns which are robust for current vision systems to recognize.
In contrast, \emph{activity attributes} %on a wide scale
are more difficult to conceptualize as they involve varying levels of abstractness, % and meaning,
which are also more challenging for computer vision as they have less distinct visual patterns.
%\citet{Antol2014} note this difficulty (for instance, of describing what it means to `sit' to a machine with no understanding of this concept) and collect data of abstract illustrations of two-person actions for generalization to dyadic activity recognition.
Noting this difficulty, \citet{Antol2014} instead employ cartoon illustrations as intermediate mappings for zero-shot dyadic activity recognition.
We present a complementary approach: that of tackling the abstractness of verb attributes directly. We develop and use a corpus of verb attributes, using linguistic theories on verb semantics (e.g., aspectual verb classes of \citet{vendler_verbs_1957}) and also drawing inspiration from studies on linguistic categorization of verbs and their properties \citep{friedrich2014automatic,siegel2000learning}.

In sum, we present the first study aiming to recover general action attributes for a diverse collection of verbs, and probe their predictive power for zero-shot activity recognition on the recently introduced imSitu dataset \cite{imsitu}.
%Our results show that learned verb attributes provide predictive signal for classifying previously unseen actions and suggest several avenues for future work on this challenging task.
Empirical results show that action attributes inferred from language can help classifying previously unseen activities and suggest several avenues for future research on this challenging task.
We publicly share our dataset and code for future research.\footnote{Available at \href{https://github.com/uwnlp/verb-attributes}{http://github.com/uwnlp/verb-attributes} }

%%%%%%%%%%%%%%%%%%%%%%%%%%%%%%%%%%%%%%%%%%%%%%%%%%%%%%%%%%%%%%%%%%%%%
\section{Action Verb Attributes}\label{sec:actionverbattributes}
We consider seven different groups of action verb attributes. They are motivated in part by potential relevance for visual zero-shot inference, and in part by classical literature on linguistic theories on verb semantics. The attribute groups are summarized below.\footnote{The full list is available in the supplemental section.} Each attribute group consists of a set of attributes, which sums to $K=24$ distinct attributes annotated over the verbs.\footnote{Several of our attributes are categorical; if converted to binary attributes, we would have 40 attributes in total.}

\paragraph{[1] Aspectual Classes}

We include the aspectual verb classes of \citet{vendler_verbs_1957}:
\begin{itemize}
\neatlist
\item \emph{state}: a verb that does not describe a changing situation (e.g. ``have'', ``be'')
\item \emph{achievement}: a verb that can be \emph{completed} in a short period of time (e.g. ``open'', ``jump'')
\item \emph{accomplishment}: a verb with a sense of completion over a longer period of time (e.g. ``climb'')
\item \emph{activity}: a verb without a clear sense of completion (e.g. ``swim'', ``walk'', ``talk'')
\end{itemize}

\paragraph{[2] Temporal Duration}
This attribute group relates to the aspectual classes above, but provides additional estimation of typical time duration with four categories. We categorize verbs by best-matching temporal units: {seconds}, minutes, hours, or days, with an additional option for verbs with unclear duration (e.g., ``provide'').
\paragraph{[3] Motion Dynamics}
This attribute group focuses on the energy level of motion dynamics in four categories: no motion (``sleep''), low motion (``smile''), medium  (``walk''), or high  (``run''). We add an additional option for verbs whose motion level depends highly on context, such as ``finish.''
\paragraph{[4] Social Dynamics}
This attribute group focuses on the likely social dynamics, in particular, whether the action is usually performed as a solitary act, a social act, or either. This is graded on a 5-part scale from least social $(-2)$ to either $(+0)$ to most social $(+2)$
\paragraph{[5] Transitivity}
This attribute group focuses on whether the verb can take an object, or be used without. This gives the model a sense of the implied action dynamics of the verb between the agent and the world. We record three variables: whether or not the verb is naturally transitive on a person (``I hug her'' is natural), on a thing (``I eat it''), and whether the verb is intransitive (``I run''). It should be noted that a small minority of verbs do not allow a person as an agent (``snow'').
\paragraph{[6] Effects on Arguments}
This attribute group focuses on the effects of actions on agents and other arguments. For each of the possible transitivities of the verb, we annotate whether or not it involves \emph{location change} (``travel''), \emph{world change} (``spill''), \emph{agent or object change} (``cry'') , or \emph{no visible change} (``ponder'').
\paragraph{[7] Body Involvements} This attribute group specifies prominent body parts involved in carrying out the action. For example, ``open'' typically involves ``hands'' and ``arms'' when used in a physical sense. We use five categories: head, arms, torso, legs, and other body parts.
\paragraph{Action Attributes and Contextual Variations}
In general, contextual variations of action attributes are common, especially for frequently used verbs that describe everyday physical activities. For example, while ``open'' typically involves ``hands'', there are exceptions, e.g. ``open one's eyes.''
In this work, we focus on stereotypical or prominent characteristics across a range of actions that can be denoted using the same verb.
Thus, three investigation points of our work include: (1) crowd-sourcing experiments to estimate the distribution of human judgments on the prominent characteristics of everyday physical action verbs, (2) the feasibility of learning models for inferring the prominent characteristics of the everyday action verbs despite the potential noise in the human annotation, and (3) their predictive power in zero-shot action recognition despite the potential noise from contextual variations of action attributes. As we will see in Section \ref{sec:results}, our study confirms the usefulness of studying action attributes and motivates the future study in this direction.

\paragraph{Relevance to Linguistic Theories}
The key idea in our work that action verbs project certain expectations about their influence on their arguments, their pre- and post-conditions, and their implications on social dynamics, etc., relates to the original Frame theories of \citet{BakerFillmoreLowe:98}. The study of action verb attributes is also closely related to formal studies on verb categorization based on the characteristics of the actions or states that a verb typically associates with \cite{levin1993}, and cognitive linguistics literature that focus on causal structure and force dynamics of verb meanings \cite{croft2012verbs}.

%%%%%%%%%%%%%%%%%%%%%%%%%%%%%%%%%%%%%%%%%%%%%%%%%%%%%%%%%%%%%%%%%%%%%%%%%%%%%%%%%%%%%%
\section{Learning Verb Attributes from Language}\label{sec:laal}
In this section we present our models for learning verb attributes from language. We consider two complementary types of language-based input: dictionary definitions and word embeddings. The approach based on dictionary definitions resembles how people acquire the meaning of a new word from a dictionary lookup, while the approach based on word embeddings resembles how people acquire the meaning of words in context.

\paragraph{Overview} This task follows the standard supervised learning approach where the goal is to predict $K$ attributes per word in the vocabulary $\mathcal{V}$. Let $x_v \in \mathcal{X}$ represent the input representation of a word $v \in \mathcal{V}$. For instance, $x_v$ could denote a word embedding, or a definition looked up from a dictionary (modeled as a list of tokens). Our goal is to produce a model $f : \mathcal{X} \to \mathbb{R}^{d}$ that maps the input to a representation of dimension $d$. Modeling options include using pretrained word embeddings, as in Section \ref{sec:emb}, or using a sequential model to encode a dictionary, as in Section \ref{sec:gru}.

Then, the estimated probability distribution over attribute $k$ is given by:
\begin{equation}\label{eq:estattdist}
\hat{y}_{v,k} = \sigma (\mathbf{W}^{(k)}f(x_v)).
\end{equation}
If the attribute is binary, then $\mathbf{W}^{(k)}$ is a vector of dimension $d$ and $\sigma$ is the sigmoid function. Otherwise, $\mathbf{W}^{(k)}$ is of shape $d_k \times d$, where $d_k$ is the dimension of attribute $k$, and $\sigma$ is the softmax function. Let the vocabulary $\mathcal{V}$ be partitioned into sets $\mathcal{V}_{train}$ and $\mathcal{V}_{test}$; then, we train by minimizing the cross-entropy loss over $\mathcal{V}_{train}$ and report attribute-level accuracy over words in $\mathcal{V}_{test}$.

\paragraph{Connection to Learning Object Attributes}
Our task of learning verb attributes used in activity recognition is related to the task of learning object attributes, but with several key differences. \citet{al2016recovering} build the ``Class-Attribute Association Prediction'' model (CAAP) that classifies the attributes of an object class from its name. They apply it on the \emph{Animals with Attributes} dataset, which contains 50 animal classes, each described by 85 attributes \citep{lampert}. Importantly, these attributes are concrete details with semantically meaningful names such as ``has horns'' and ``is furry.'' The CAAP model takes advantage of this, consisting of a tensor factorization model initialized by the word embeddings of the object class names as well as the attribute names. On the other hand, \emph{verb attributes} such as the ones we outline in Section \ref{sec:actionverbattributes}, are technical linguistic terms. Since word embeddings principally capture common word senses, they are unsuited for verb attributes. Thus, we evaluate two versions of CAAP as a baseline: CAAP-pretrained, where where the model is preinitialized with GloVe embeddings for the attribute names \citep{pennington2014glove}, and CAAP-learned, where the model is learned from random initialization.

\subsection{Learning from Distributed Embeddings}\label{sec:emb}
One way of producing attributes is from distributed word embeddings such as word2vec \citep{mikolov2013distributed}. Intuitively, we expect similar verbs to have similar distributions of nearby nouns and adverbs, which can greatly help us in zero-shot prediction. In our experiments, we use 300-dimensional
GloVe vectors trained on 840B tokens of web data \citep{pennington2014glove}. We use logistic regression to predict each attribute, as we found that extra hidden layers did not improve performance. Thus, we let $f^{emb}(x_v) = \mathbf{w}_v$, the GloVe embedding of $v$, and use Equation~\ref{eq:estattdist} to get the distribution over labels.

We additionally experiment with retrofitted embeddings, in which embeddings are mapped in accordance with a lexical resource. Following the approach of \citet{faruqui_retrofitting_2015}, we retrofit embeddings using WordNet \citep{miller1995wordnet}, Paraphrase-DB \citep{ganitkevitch2013ppdb}, and FrameNet \citep{baker1998berkeley}.

\subsection{Learning from Dictionary Definitions}\label{sec:gru}

We additionally propose a model that learns the attribute-grounded meaning of verbs through dictionary definitions. This is similar in spirit to the task of using a dictionary to predict word embeddings  \citep{hill_learning_2016}.

\paragraph{BGRU encoder} Our first model involves a Bidirectional Gated Recurrent Unit (BGRU) encoder \citep{cho-EtAl:2014:EMNLP2014}. Let $x_{v,1:T}$ be a definition for verb $v$, with $T$ tokens. To encode the input, we pass it through the GRU equation:
\begin{equation}\label{eq:gru}
\vec{\mathbf{h}}_t = \textrm{GRU}(x_{v,t}, \vec{\mathbf{h}}_{t-1}).
\end{equation}
% \begin{bmatrix} \mathbf{r} \\ \mathbf{z} \end{bmatrix} =
%  \sigma \left( \begin{bmatrix} \mathbf{W}_r \\ \mathbf{W}_z \end{bmatrix}\mathbf{x}_{v, t-1} \begin{bmatrix} \mathbf{U}_r \\ \mathbf{U}_z \end{bmatrix} \mathbf{h}_{t-1}\right)
Let $\cev{\mathbf{h}}_1$ denote the output of a GRU applied on the reversed input $x_{v,T:1}$. Then, the BGRU encoder is the concatenation $f^{bgru} =  \vec{\mathbf{h}}_T\| \cev{\mathbf{h}}_1$.

\paragraph{Bag-of-words encoder}
Additionally, we try two common flavors of a Bag-of-Words model. In the standard case, we first construct a vocabulary of 5000 words by frequency on the dictionary definitions. Then, $f^{bow}(x_v)$ represents the one-hot encoding $f^{bow}(x_v)_i = \left[i \in x_v\right]$, in other words, whether word $i$ appears in definiton $x_v$ for verb $v$.

Additionally, we try out a Neural Bag-of-Words model where the GloVe embeddings in a definition are averaged \citep{iyyer2015deep}. This is $f^{nbow}(x_{v,1:T}) = \frac{1}{|T|}\sum_{t=1}^T f^{emb}(x_{v,t})$.
%\begin{equation}\label{eq:nbow}
%f^{nbow}(x_{v,1:T}) = \frac{1}{|T|}\sum_{t=1}^T f^{emb}(x_{v,t})
%\end{equation}
\begin{figure}[t]
\centerline{\includegraphics[scale=0.7]{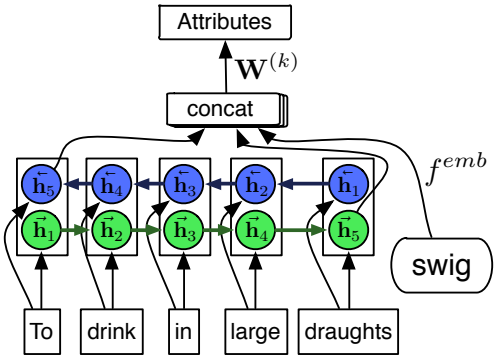}}
\caption{Overview of our combined dictionary + embedding to attribute model. Our encoding is the concatenation of a Bidirectional GRU of a definition and the word embedding for that word. The encoding is then mapped to the space of attributes using parameters $\mathbf{W}^{(k)}$.}
\label{fig:dict}
\end{figure}
\paragraph{Dealing with multiple definitions per verb}
One potential pitfall with using dictionary definitions is that there are often many defnitions associated with each verb. This creates a dataset bias since polysemic verbs are seen more often. Additionally, dictionary definitions tend to be sorted by relevance, thus lowering the quality of the data if all definitions are weighted equally during training. To counteract this, we randomly oversample the definitions at training time so that each verb has the same number of definitions.\footnote{For the (neural) bag of words models, we also tried concatenating the definitions together per verb and then doing the encoding. However, we found that this gave worse results.} At test time, we use the first-occurring (and thus generally most relevant) definition per verb.

\subsection{Combining Dictionary and Embedding Representations}
We hypothesize that the two modalities of the dictionary and distributional embeddings are complementary. Therefore, we propose an early fusion (concatenation) of both categories. Figure~\ref{fig:dict} describes the GRU + embedding model--in other words, $f^{BGRU+emb} = f^{BGRU} \| f^{emb}$. This can likewise be done with any choice of definition encoder and word embedding.

%%%%%%%%%%%%%%%%%%%%%%%%%%%%%%%%%%%%%%%%%%%%%%%%%%%%%%%%%%%%%%%%%%%%%%%%%%%%%%
\section{Zero-Shot Activity Recognition}
\label{sec:zsl}
%%%%%%%%%%%%%%%%%%%%%%%%%%%%%%%%%%%%%%%%%%%%%%%%%%%%%%%%%%%%%%%%%%%%%%%%%%%%%%

%%%%%%%%%%%%%%%%%%%%%%%%%%%%%%%%%%%%%%%%%%%%%%%%%%%%%%%%%%%%%%%%%%%%%%%%%%%%%%
\subsection{Verb Attributes as Latent Mappings}
\label{sec:att-zsl}
%%%%%%%%%%%%%%%%%%%%%%%%%%%%%%%%%%%%%%%%%%%%%%%%%%%%%%%%%%%%%%%%%%%%%%%%%%%%%%
Given learned attributes for a collection of activities, we would like to evaluate their performance at describing these activities from real world images in a zero-shot setting. Thus, we consider several models that classify an image's label by pivoting through an attribute representation.

\paragraph{Overview} A formal description of the task is as follows. Let the space of labels be $\mathcal{V}$, partitioned into $\mathcal{V}_{train}$ and $\mathcal{V}_{test}$. Let $z_v \in \mathcal{Z}$ represent an image with label $v \in \mathcal{V}$; our goal is to correctly predict this label amongst verbs $v \in \mathcal{V}_{test}$ at test time, despite never seeing any images with labels in $\mathcal{V}_{test}$ during training.

Generalization will be done through a lookup table $\mathbf{A}$, with known attributes for each $v \in \mathcal{V}$. Formally, for each attribute $k$ we define it as:
\begin{equation} \label{eq:atttable}
\mathbf{A}_{v', i}^{(k)} = \begin{cases}
+1 & \textrm{if attribute $k$ for verb $v'$ is $i$} \\
-1 & \textrm{otherwise}
\end{cases}
\end{equation}
For binary attributes, we need only one entry per verb, making $\mathbf{A}^{(k)}$ a single column vector.
%Let $\mathbf{A}_{v',k}$ denote the value of the $k$th attribute for a verb $v'$: if attribute $k$ is binary, then $\mathbf{A}_{v',k} \in \{-1, 1\}$, otherwise, if attribute $k$ is categorical, then $\mathbf{A}_{v',k}$ is a one-hot encoding.
Let our image encoder be represented by the map $g : \mathcal{Z} \to \mathbb{R}^{d}$. We then use the linear map in Equation~\ref{eq:estattdist} to produce the log-probability distribution over each attribute $k$. The distribution over labels is then:
\begin{equation}\label{eq:attprob}
P(\cdot| z_v) = \underset{v'}{\textrm{softmax}} \left( \sum_k \mathbf{A}^{(k)} \mathbf{W}^{(k)} g(z_v) \right)
\end{equation}
where $\mathbf{W}^{(k)}$ is a learned parameter that maps the image encoder to the attribute representation. We then train our model by minimizing the cross-entropy loss over the training verbs $\mathcal{V}_{train}$.

\paragraph{Convolutional Neural Network (CNN) encoder}
Our image encoder is a CNN with the ResNet-152 architecture \citep{he2016deep}. We use weights pretrained on ImageNet \citep{imagenet_cvpr09} and perform additional pretraining on imSitu using the classes $\mathcal{V}_{train}$. After this, we remove the top layer and set $g(x_v)$ to be the 2048-dimensional image representation from the network.

\subsubsection{Connection to Other Attribute Models} \label{sec:otherzsl}
Our model is similar to those %the zero-shot models
of \citet{akata2013label} and \mbox{\citet{romera2015embarrassingly}} in that we predict the attributes indirectly and train the model through the class labels.\footnote{Unlike these models, however, we utilize (some) categorical attributes and optimize using cross-entropy.} It differs from several other zero-shot models, such as \citet{lampert}'s Direct Attribute Prediction (DAP) model, in that DAP is trained by maximizing the probability of predicting each attribute and then multiplying the probabilities at test time. Our use of joint training the recognition model to directly optimize class-discrimination rather than attribute-level accuracy.

\subsection{Verb Embeddings as Latent Mappings}
\label{sec:emb-zsl}
An additional method of doing zero-shot image classification is by using word embeddings directly. \citet{frome_devise:_2013} build DeVISE, a model for zero-shot learning on ImageNet object recognition where the objective is for the image model to predict a class's word embedding directly. DeVISE is trained by minimizing
\begin{equation}\label{eq:devise}\nonumber
\sum_{v'\in \mathcal{V}_{train} \setminus \{v\}} \max\{0, .1 + (\mathbf{w}_{v'}-\mathbf{w}_v)\mathbf{W}^{emb}g(z_v) \}
\end{equation}
for each image $z_v$. We compare against a version of this model with fixed GloVe embeddings $\mathbf{w}$.

Additionally, we employ a variant of our model using only word embeddings. The equation is the same as Equation~\ref{eq:attprob}, except using the matrix $\mathbf{A}^{emb}$ as a matrix of word embeddings: i.e., for each label $v$ we consider, we have $\mathbf{A}^{emb}_{v} = \mathbf{w}_{v}.$

\subsection{Joint Prediction from Attributes and Embeddings}
\begin{figure}[t]
\centerline{\includegraphics[scale=1]{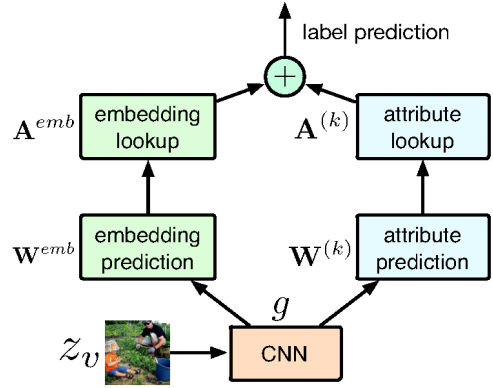}}
\vspace{-1mm}
\caption{Our proposed model for combining attribute-based zero-shot learning and word-embedding based transfer learning. The embedding and attribute lookup layers are used to predict a distribution of labels over $\mathcal{V}_{train}$ during training and $\mathcal{V}_{test}$ during testing.}
\label{fig:ensemble}
\end{figure}

To combine the representation power of the attribute and embedding models, we build an ensemble combining both models. This is done by adding the logits before the softmax is applied: \[
\underset{v'}{\textrm{softmax}} \mkern-4mu\left(\mkern-4mu\sum_k \mathbf{A}^{(k)}\mathbf{W}^{(k)}g(z_v) + \mathbf{A}^{emb}\mathbf{W}^{emb}g(z_v)\mkern-4mu\right)
\]
A diagram is shown in Figure~\ref{fig:ensemble}. We find that during optimization, this model can easily overfit, presumably by excessive coadaption of the embedding and attribute components. To solve this, we train the model to minimize the cross entropy of three sources independently: the attributes only, the embeddings only, and the sum, weighting each equally.

\paragraph{Incorporating predicted and gold attributes}
We additionally experiment with an ensemble of our model, combining predicted and gold attributes of $\mathcal{V}_{test}$. This allows the model to hedge against cases where a verb attribute might have several possible correct answers.  A single model is trained; at test time, we multiply the class level probabilities $P(\cdot | z_v)$ of each together to get the final predictions.

%%%%%%%%%%%%%%%%%%%%%%%%%%%%%%%%%%%%%%%%%%%%%%%%%%%%%%%%%%%%%%%%%%%%%%%% TEXT RESULTS
\begin{table*}[t!]
\hspace*{-4mm}
\centering
\small
\begin{tabular}{  l l || r  r | r  r  r  r  r  r  r}
 &                      & acc-macro & acc-micro & Body & Duration & Aspect & Motion & Social & Effect & Transi. \\ \hline
 & most frequent class & 61.33 & 75.45 & 76.84 & 76.58 & 43.67 & 35.13 & 42.41 & 84.97 & 69.73 \\ \hline
\multirow{6}{*}{\rotatebox[origin=c]{90}{Emb}}
& CAAP-pretrained & 64.96 & 78.06 & 84.81 & 72.15 & 50.95 & 46.84 & 43.67 & 85.21 & 71.10 \\
& CAAP-learned & 68.15 & 81.00 & 86.33 & 76.27 & 52.85 & 52.53 & \bf{45.57} & 88.29 & 75.21 \\
& GloVe                                & 66.60 & 79.69 & 85.76 & 75.00 & 50.32 & 48.73 & 43.99 & 86.52 & 75.84 \\
& GloVe + framenet                       & 67.42 & 80.79 & 86.27 & 76.58 & 49.68 & 50.32 & 44.94 & 88.19 & 75.95 \\
& GloVe + ppdb                           & 67.52 & 80.75 & 85.89 & 76.58 & 51.27 & 50.95 & 43.99 & 88.21 & 75.74 \\
& GloVe + wordnet                        & 68.04 & 81.13 & 86.58 & 76.90 & 54.11 & 50.95 & 43.04 & \bf{88.34} & \bf{76.37} \\ \hline
\multirow{3}{*}{\rotatebox[origin=c]{90}{Dict}}
 &  BGRU                              & 66.05 & 79.44 & 85.70 & 76.90 & 51.27 & 48.42 & 40.51 & 86.92 & 72.68 \\
 & BoW                                 & 62.53 & 77.61 & 83.54 & 76.58 & 48.42 & 35.76 & 36.39 & 86.31 & 70.68 \\
 & NBoW                                & 65.41 & 78.96 & 85.00 & 76.58 & 52.85 & 42.41 & 43.35 & 86.87 & 70.78 \\ \hline
\multirow{3}{*}{\rotatebox[origin=c]{90}{D+E}}
& NBoW + GloVe                         & 67.52 & 80.76 & \bf{86.84} & 75.63 & 53.48 & 51.90 & 41.77 & 88.03 & 75.00 \\
& BoW + GloVe                          & 63.15 & 77.89 & 84.11 & \bf{77.22} & 49.68 & 34.81 & 38.61 & 86.18 & 71.41 \\
 &  BGRU + GloVe                      & \bf{68.43} & \bf{81.18} & 86.52 & 76.58 & \bf{56.65} & \bf{53.48} & 41.14 & 88.24 & \bf{76.37} \\
\end{tabular}
\caption{Results on the text-to-attributes task. All values reported are accuracies (in \%). For attributes where multiple labels can be selected, the accuracy is averaged over all instances (e.g., the accuracy of ``Body'' is given by the average of accuracies from correctly predicting Head, Torso, etc.). As such, we report two ways of averaging the results: microaveraging (where the accuracy is the average of accuracies on the underlying labels) and macroaveraging (where the accuracy is averaged together from the groups).
}
\label{text-results}
\end{table*}
%%%%%%%%%%%%%%%%%%%%%%%%%%%%%%%%%%%%%%%%%%%%%%%%%%%%%%%%%%%%%%%%%%%%%%%%%%%%%%%%%%%%%%%%%%%%%%%%%%%%%%%%%%%%%%%%%%%%%%%%%%%%%%%%%

\section{Actions and Attributes Dataset}

To evaluate our hypotheses on action attributes and zero-shot learning, we constructed a dataset using crowd-sourcing experiments.  The \emph{Actions and Attributes} dataset consists of annotations for 1710 verb templates, each consisting of a verb and an optional particle (e.g. ``put'' or ``put \emph{up}'').%\footnote{Available at \mbox{\href{https://github.com/rowanz/verb-attributes}{github.com/rowanz/verb-attributes}} }

We selected all verbs from the imSitu corpus, which consists of images representing verbs from many categories \citep{imsitu}, then extended the set using the MPII movie visual description dataset and ScriptBase datasets, \citep{rohrbach_dataset_2015,gorinski-lapata:2015:NAACL-HLT}. We used the spaCy dependency parser \citep{honnibal-johnson:2015:EMNLP} to extract the verb template for each sentence, and collected annotations on Mechanical Turk to filter out nonliteral and abstract verbs. Turkers annotated this filtered set of templates using the attributes described in Section \ref{sec:actionverbattributes}. In total, 1203 distinct verbs are included. The templates are split randomly by verb; out of 1710 total templates, we save 1313 for training, 81 for validation, and 316 for testing.

To provide signal for classifying these verbs, we collected dictionary definitions for each verb using the Wordnik API,\footnote{Available at \url{http://developer.wordnik.com/} with access to American Heriatge Dictionary, the Century Dictionary, the GNU Collaborative International Dictionary, Wordnet, and Wiktionary.} including only senses that are explicitly labeled ``verb.'' This leaves us with 23,636 definitions, an average of 13.8 per verb.

\section{Experimental Setup}
\paragraph{BGRU pretaining} We pretrain the BGRU model on the Dictionary Challenge, a collection of 800,000 word-definition pairs obtained from Wordnik and Wikipedia articles \citep{hill_learning_2016}; the objective is to obtain a word's embedding given one of its definitions. For the BGRU model, we use an internal dimension of 300, and embed the words to a size 300 representation. The vocabulary size is set to 30,000 (including all verbs for which we have definitions).  During pretraining, we keep the architecture the same, except a different 300-dimensional final layer is used to predict the GloVe embeddings.

Following \citet{hill_learning_2016}, we use a ranking loss. Let $\hat{\mathbf{w}} = \mathbf{W}^{emb}f(x)$ be the predicted word embeddings for each definition $x$ of a word in the dictionary (not necessarily a verb). Let $\mathbf{w}$ be the word's embedding, and $\widetilde{\mathbf{w}}$ be the embedding of a random dictionary word. The loss is then given by:
\vspace*{-3mm}
\begin{equation}
\label{eq:marginloss}
L = \max\{0, .1- \cos(\mathbf{w}, \hat{\mathbf{w}}) + \cos(\mathbf{w}, \widetilde{\mathbf{w}})\}\nonumber
\end{equation}
After pretraining the model on the Dictionary Challenge, we fine-tune the attribute weights $\mathbf{W}^{(k)}$ using the cross-entropy over Equation~\ref{eq:estattdist}.

\paragraph{Zero-shot with the imSitu dataset}
We build our image-to-verb model on the imSitu dataset, which contains a diverse collection of images depicting one of 504 verbs. The images represent a variety of different semantic role labels \cite{imsitu}. Figure~\ref{vis-examples} shows examples from the dataset. We apply our attribute split to the dataset and are left with 379 training classes, 29 validation classes, and 96 test classes.

\paragraph{Zero-shot activity recognition baselines}
We compare against several additional baseline models for learning from attributes and embeddings. \mbox{\citet{romera2015embarrassingly}}
propose ``Embarassingly Simple Zero-shot Learning'' (ESZL), a linear model that directly predicts class labels through attributes and incorporates several types of regularization. We compare against a variant of \mbox{\citet{lampert}}'s DAP model discussed in Section~\ref{sec:otherzsl}. We additionally compare against DeVISE \citep{frome_devise:_2013}, as mentioned in Section~\ref{sec:emb-zsl}. We use a ResNet-152 CNN finetuned on the imSitu $\mathcal{V}_{train}$ classes as the visual features for these baselines (the same as discussed in Section \ref{sec:att-zsl}).

\paragraph{Additional implementation details} are provided in the Appendix.
%%%%%%%%%%%%%%%%%%%%%%%%%%%%%%%%%%%%%%%%%%%%%%%%%%%%%%%%%%%%%%%%%%%%%%%%%%%%%%%%%%%%%%%%%%%%%%%%%%%%%%%%%%%%%%%%%%%%%%%%%%%%%%%%%

\section{Experimental Results} \label{sec:results}

\subsection{Predicting Action Attributes from Text}
Our results for action attribute prediction from text are given in Table~\ref{text-results}. Several examples are given in the supplemental section in Table~\ref{text-examples}. Our results on the text-to-attributes challenge confirm that it is a challenging task for two reasons. First, there is noise associated with the attributes: many verb attributes are hard to annotate given that verb meanings can change in context.\footnote{ As such, our attributes have a median Krippendorff Alpha of $\alpha=.359$.} Second, there is a lack of training data inherent to the problem: there are not many common verbs in English, and it can be difficult to crowdsource annotations for rare ones. Third, any system must compete with strong frequency-based baselines, as attributes are generally sparse. Moreover, we suspect that were more attributes collected (so as to cover more obscure patterns), the sparsity would only increase.

Despite this, we report strong baseline results on this problem, particularly with our embedding based models. The performance gap between embedding-only and definition-only models can possibly be explained by the fact that the word embeddings are trained on a very large corpus of real-world \emph{examples} of the verb, while the definition is only a single high-level representation meant to be understood by someone who already speaks that language. For instance, it is likely difficult for the definition-only model to infer whether a verb is transitive or not (Transi.), since definitions might assume commonsense knowledge about the underlying concepts the verb represents. The strong performance of embedding models is further enhanced by using retrofitted word embeddings, suggesting an avenue for improvement on language grounding through better representation of linguistic corpora.

We additionally see that both joint dictionary-embedding models outperform the dictionary-only models overall. In particular, the BGRU+GloVe model performs especially well at determining the aspect and motion attributes of verbs, particularly relative to the baseline. The strong performance of the BGRU+GloVe model indicates that there is some signal that is missing from the distributional embeddings that can be recovered from the dictionary definition. We thus use the predictions of this model for zero-shot image recognition.

Based on error analysis, we found that one common mode of failure is where commonsense knowledge is required. To give an example, the embedding based model labels ``shop'' as a likely solitary action. This is possibly because there are a lack of similar verbs in $\mathcal{V}_{train}$; by random chance, ``buy'' is also in the test set. We see that this can be partially mitigated by the dictionary, as evidenced by the fact that the dictionary-based models label ``shop'' as in between social and solitary. Still, it is a difficult task to infer that people like to ``visit stores in search of merchandise'' together.
%%%%%%%%%%%%%%%%%%%%%%%%%%%%%%%%%%%%%%%%%%%%%%%%%%%%%%%%%%%%%%%%%%%%%% OTHER RESULTS
\subsection{Zero-shot Action Recognition}
\begin{table}[!t]
\centering
\begin{tabular}{@{\hskip0pt}r@{\hskip5pt} | @{\hskip2pt} c @{\hskip2pt}  @{\hskip2pt} c @{\hskip2pt}  @{\hskip2pt} c @{\hskip3pt} ||  r @{\hskip5pt} r @{\hskip0pt}}
  &  \multicolumn{3}{c||}{Attributes used} &  \multicolumn{2}{c}{$v \in \mathcal{V}_{test}$}  \\
 Model & atts(P) & atts(G) & GloVe & top-1 & top-5 \\ \hline \hline
 Random & & &  & 1.04 & 5.20 \\ \hline
 %GVLE & & & \checkmark    & 1.04 & 5.21 \\ \hline
 DeVISE & & & \checkmark  & 16.50&37.56  \\ \hline
\multirow{2}{*}{ESZL} & & \checkmark &  & 3.60 & 14.81 \\
 & \checkmark & &  & 3.27 & 13.21 \\ \hline
\multirow{2}{*}{DAP}    & & \checkmark &  & 3.35&16.69 \\
                          & \checkmark & &  & 4.33 & 17.56 \\ \hline
 \multirow{7}{*}{Ours}
   &  & \checkmark &  & 4.79&19.98  \\
   & \checkmark & &  & 7.04&22.19 \\
   & \checkmark & \checkmark & & 7.71&24.90 \\
   &  &  & \checkmark & 17.60&39.29 \\
   & & \checkmark &  \checkmark & 18.10&41.46  \\
   & \checkmark &  & \checkmark & 16.75&40.44  \\
  & \checkmark & \checkmark & \checkmark & \bf{18.15} &\bf{42.17} \\
\end{tabular}
\caption{Results on the image-to-verb task. atts(P) refers to attributes predicted from the BGRU+GloVe model described in Section~\ref{sec:laal}, atts(G) to gold attributes, and GloVe to GloVe vectors. The accuracies reported are amongst the 96 unseen labels of $\mathcal{V}_{test}$.}
\label{img-results}
\end{table}

Our results for verb prediction from images are given in Table~\ref{img-results}. Despite the difficulty of predicting the correct label over 96 unseen choices, our models show predictive power. Although our attribute models do not outperform our embedding models and DeVISE alone,  we note that our joint attribute and embedding model scores the best overall, reaching 18.10\% in top-1 and 41.46\% in top-5 accuracy when using gold attribute annotations for the zero-shot verbs. This result is possibly surprising given the small number of attributes ($K=24$) in total, of which most tend to be sparse (as can be seen from the baseline performance in Table~\ref{text-results}). We thus hypothesize that collecting more activity attributes would further improve performance.

\begin{figure*}[t]
\centerline{\includegraphics[scale=0.56]{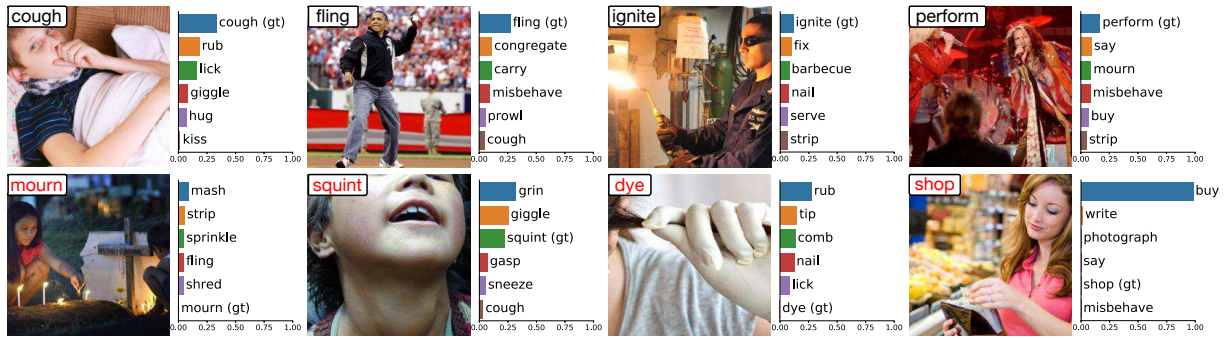}}
\vspace{-1mm}
\caption{Predictions on unseen imSitu classes from our attribute+embedding model with gold attributes. The top and bottom rows show success and failure cases, respectively. The bars to the right of each image represent a probability distribution, showing the ground truth class and the top 5 scoring incorrect classes.
}
\label{vis-examples}
\vspace{-0mm}
\end{figure*}

We also note the success in performing zero-shot learning with predicted attributes. Perhaps paradoxically, our attribute-only models (along with DAP) perform better in both accuracy metrics when given predicted attributes at test time, as opposed to gold attributes. Further, we get an extra boost by ensembling predictions of our model when given two sets of attributes at test time, giving us the best results overall at 18.15\% top-1 accuracy and 42.17\% top-5. Interestingly, better performance with predicted attributes is also reported by \citet{al2016recovering}: predicting the attributes with their CAAP model and then running the DAP model on these predicted attributes outperforms the use of gold attributes at test time. It is somewhat unclear why this is the case--possibly, there is some bias in the attribute labeling, which the attribute predictor can correct for.

In addition to quantitative results, we show some zero-shot examples in Figure~\ref{vis-examples}. The examples show inherent difficulty of zero-shot action recognition. Incorrect predictions are often reasonably related to the situation (``rub'' vs ``dye'') but picking the correct target verb based on attribute-based inference is still a challenging task.

Although our results appear promising, we argue that our model still fails to represent much of the semantic information about each image class. In particular, our model is prone to \emph{hubness}: the overprediction of a limited set of labels at test time (those that closely match signatures of examples in the training set). This problem has previously been observed with the use of word embeddings for zero-shot learning \citep{marco2015hubness} and can be seen in our examples (for instance, the over-prediction of ``buy''). Unfortunately, we were unable to mitigate this problem in a way that also led to better quantitative results (for instance, by using a ranking loss as in DeVISE \citep{frome_devise:_2013}). We thus leave resolving the hubness problem in zero-shot activity recognition as a question for future work.

%%%%%%%%%%%%%%%%%%%%%%%%%%%%%%%%%%%%%%%%%%%%%%%%%%%
\section{Related Work}
\label{sec:related}
%%%%%%%%%%%%%%%%%%%%%%%%%%%%%%%%%%%%%%%%%%%%%%%%%%%
\paragraph{Learning attributes from embeddings}
\mbox{\citet{rubinstein-levy-schwartz-rappoport:2015:acl}} seek to predict \citet{McRae2005}'s feature norms from word embeddings of concrete nouns. Likewise, the CAAP model of \citet{al2016recovering}
predicts the object attributes of concrete nouns for use in zero-shot learning. In contrast, we predict verb attributes. A related task is that of improving word embeddings using multimodal data and linguistic resources \citep{faruqui_retrofitting_2015,silberer-ferrari-lapata:2013:ACL2013,VendrovArxiv15}. Our work runs orthogonal to this, as we focus on word attributes as a tool for a zero-shot activity recognition pipeline.
\paragraph{Zero-shot learning with objects}
Though distinct, our work is related to zero-shot learning of objects in computer vision. There are several datasets \citep{Nilsback08,welinder2010caltech}
and models developed on this task (\mbox{\citet{romera2015embarrassingly}}; \citet{lampert}; \citet{mukherjee2016gaussian}; \mbox{\citet{farhadi2010attribute}}). In addition, \mbox{\citet{BaICCV15}} augment existing datasets with descriptive Wikipedia articles so as to learn novel objects from descriptive text.
As illustrated in Section~\ref{sec:intro}, action attributes pose unique challenges compared to object attributes, thus models developed for zero-shot object recognition are not as effective for zero-shot action recognition, as has been empirically shown in Section~\ref{sec:results}.

\paragraph{Zero-shot activity recognition}In prior work, zero-shot activity recognition has been studied on video datasets, each containing a selection of concrete physical actions. The MIXED action dataset, itself a combination of three action recognition datasets, has 2910 labeled videos with 21 actions, each described by 34 action attributes \citep{liu2011recognizing}. These action attributes are concrete binary attributes corresponding to low-level physical movements, for instance, ``arm only motion,'' ``leg: up-forward motion.'' By using word embeddings instead of attributes, \citet{xu2017transductive} study video activity recognition on a variety of action datasets, albeit in the transductive setting wherein access to the test labels is provided during training. In comparison with our work on imSitu, these video datasets lack broad coverage of verb-level classes (and for some, sufficient data points per class).

The abstractness of broad-coverage activity labels makes the problem much more difficult to study with attributes. To get around this, \citet{Antol2014} present a synthetic dataset of cartoon characters performing dyadic actions, and use these cartoon illustrations as internal mappings for zero-shot recognition of dyadic actions in real-world images.
We investigate an alternative approach
by using linguistically informed verb attributes for activity recognition.

\section{Future Work and Conclusion}
Several possibilities remain open for future work. First, more attributes could be collected and evaluated, possibly integrating other linguistic theories about verbs. %While this would likely improve the results, there are other more significant changes to our experimental setup that could be considered.
Second, future work could move beyond using attributes as a pivot between language and vision domains. In particular, since our experiments show that unsupervised word embeddings significantly help performance, it might be desirable to learn data-driven attributes in an end-to-end fashion directly from a large corpus or from dictionary definitions.
Third, future research on action attributes should ideally include videos to better capture attributes that require temporal signals.

Overall, however, our work presents a strong early step towards zero-shot activity recognition, a relatively less studied task that poses several unique challenges over zero-shot object recognition. We  introduce new action attributes motivated by linguistic theories and demonstrate their empirical use for reasoning about previously unseen activities.

\paragraph{Acknowledgements}
We thank the anonymous reviewers along with Mark Yatskar, Luke Zettlemoyer, Yonatan Bisk, Maxwell Forbes, Roy Schwartz, and Mirella Lapata, for their helpful feedback. We also thank the Mechanical Turk workers and members of the XLab, who helped with the annotation process. This work is supported by the National Science Foundataion Graduate Research Fellowship (DGE-1256082), the NSF grant (IIS-1524371), DARPA CwC program through ARO (W911NF-15-1-0543), and gifts by Google and Facebook.

\bibliography{emnlp2017}

\begin{thebibliography}{41}
\expandafter\ifx\csname natexlab\endcsname\relax\def\natexlab#1{#1}\fi

\bibitem[{Akata et~al.(2013)Akata, Perronnin, Harchaoui, and
  Schmid}]{akata2013label}
Zeynep Akata, Florent Perronnin, Zaid Harchaoui, and Cordelia Schmid. 2013.
\newblock Label-embedding for attribute-based classification.
\newblock In \emph{Proceedings of the IEEE Conference on Computer Vision and
  Pattern Recognition}, pages 819--826.

\bibitem[{Al-Halah et~al.(2016)Al-Halah, Tapaswi, and
  Stiefelhagen}]{al2016recovering}
Ziad Al-Halah, Makarand Tapaswi, and Rainer Stiefelhagen. 2016.
\newblock Recovering the missing link: Predicting class-attribute associations
  for unsupervised zero-shot learning.
\newblock In \emph{Proceedings of the IEEE Conference on Computer Vision and
  Pattern Recognition}, pages 5975--5984.

\bibitem[{Antol et~al.(2014)Antol, Zitnick, and Parikh}]{Antol2014}
Stanislaw Antol, C.~Lawrence Zitnick, and Devi Parikh. 2014.
\newblock {Zero-Shot Learning via Visual Abstraction}.
\newblock In \emph{ECCV}.

\bibitem[{Ba et~al.(2015)Ba, Swersky, Fidler, and Salakhutdinov}]{BaICCV15}
Jimmy Ba, Kevin Swersky, Sanja Fidler, and Ruslan Salakhutdinov. 2015.
\newblock Predicting deep zero-shot convolutional neural networks using textual
  descriptions.
\newblock In \emph{ICCV}.

\bibitem[{Baker et~al.(1998{\natexlab{a}})Baker, Fillmore, and
  Lowe}]{BakerFillmoreLowe:98}
Collin~F. Baker, Charles~J. Fillmore, and John~B. Lowe. 1998{\natexlab{a}}.
\newblock The {B}erkeley {F}rame{N}et project.
\newblock In \emph{COLING-ACL {\textquoteright}98: Proceedings of the
  Conference}, pages 86--90, Montreal, Canada.

\bibitem[{Baker et~al.(1998{\natexlab{b}})Baker, Fillmore, and
  Lowe}]{baker1998berkeley}
Collin~F Baker, Charles~J Fillmore, and John~B Lowe. 1998{\natexlab{b}}.
\newblock The berkeley framenet project.
\newblock In \emph{Proceedings of the 36th Annual Meeting of the Association
  for Computational Linguistics and 17th International Conference on
  Computational Linguistics-Volume 1}, pages 86--90. Association for
  Computational Linguistics.

\bibitem[{Cho et~al.(2014)Cho, van Merrienboer, Gulcehre, Bahdanau, Bougares,
  Schwenk, and Bengio}]{cho-EtAl:2014:EMNLP2014}
Kyunghyun Cho, Bart van Merrienboer, Caglar Gulcehre, Dzmitry Bahdanau, Fethi
  Bougares, Holger Schwenk, and Yoshua Bengio. 2014.
\newblock \href {http://www.aclweb.org/anthology/D14-1179} {Learning phrase
  representations using rnn encoder--decoder for statistical machine
  translation}.
\newblock In \emph{Proceedings of the 2014 Conference on Empirical Methods in
  Natural Language Processing (EMNLP)}, pages 1724--1734, Doha, Qatar.
  Association for Computational Linguistics.

\bibitem[{Croft(2012)}]{croft2012verbs}
William Croft. 2012.
\newblock \emph{Verbs: Aspect and causal structure}.
\newblock OUP Oxford.
\newblock Pg. 16.

\bibitem[{Deng et~al.(2009)Deng, Dong, Socher, Li, Li, and
  Fei-Fei}]{imagenet_cvpr09}
Jia Deng, Wei Dong, Richard Socher, Li-Jia Li, Kai Li, and Li~Fei-Fei. 2009.
\newblock Imagenet: A large-scale hierarchical image database.
\newblock In \emph{Computer Vision and Pattern Recognition, 2009. CVPR 2009.
  IEEE Conference on}, pages 248--255. IEEE.

\bibitem[{Farhadi et~al.(2010)Farhadi, Endres, and
  Hoiem}]{farhadi2010attribute}
Ali Farhadi, Ian Endres, and Derek Hoiem. 2010.
\newblock Attribute-centric recognition for cross-category generalization.
\newblock In \emph{Computer Vision and Pattern Recognition (CVPR), 2010 IEEE
  Conference on}, pages 2352--2359. IEEE.

\bibitem[{Faruqui et~al.(2015)Faruqui, Dodge, Jauhar, Dyer, Hovy, and
  Smith}]{faruqui_retrofitting_2015}
Manaal Faruqui, Jesse Dodge, Sujay~Kumar Jauhar, Chris Dyer, Eduard Hovy, and
  Noah~A Smith. 2015.
\newblock Retrofitting {Word} {Vectors} to {Semantic} {Lexicons}.
\newblock Association for Computational Linguistics.

\bibitem[{Friedrich and Palmer(2014)}]{friedrich2014automatic}
Annemarie Friedrich and Alexis Palmer. 2014.
\newblock Automatic prediction of aspectual class of verbs in context.
\newblock In \emph{ACL (2)}, pages 517--523.

\bibitem[{Frome et~al.(2013)Frome, Corrado, Shlens, Bengio, Dean, Ranzato, and
  Mikolov}]{frome_devise:_2013}
Andrea Frome, Greg~S Corrado, Jon Shlens, Samy Bengio, Jeff Dean, Marco~Aurelio
  Ranzato, and Tomas Mikolov. 2013.
\newblock \href
  {http://papers.nips.cc/paper/5204-devise-a-deep-visual-semantic-embedding-model.pdf}
  {{DeViSE}: {A} {Deep} {Visual}-{Semantic} {Embedding} {Model}}.
\newblock In C.~J.~C. Burges, L.~Bottou, M.~Welling, Z.~Ghahramani, and K.~Q.
  Weinberger, editors, \emph{Advances in {Neural} {Information} {Processing}
  {Systems} 26}, pages 2121--2129. Curran Associates, Inc.

\bibitem[{Ganitkevitch et~al.(2013)Ganitkevitch, Van~Durme, and
  Callison-Burch}]{ganitkevitch2013ppdb}
Juri Ganitkevitch, Benjamin Van~Durme, and Chris Callison-Burch. 2013.
\newblock Ppdb: The paraphrase database.
\newblock In \emph{HLT-NAACL}, pages 758--764.

\bibitem[{Gorinski and Lapata(2015)}]{gorinski-lapata:2015:NAACL-HLT}
Philip~John Gorinski and Mirella Lapata. 2015.
\newblock \href {http://www.aclweb.org/anthology/N15-1113} {Movie script
  summarization as graph-based scene extraction}.
\newblock In \emph{Proceedings of the 2015 Conference of the North American
  Chapter of the Association for Computational Linguistics: Human Language
  Technologies}, pages 1066--1076, Denver, Colorado. Association for
  Computational Linguistics.

\bibitem[{He et~al.(2016)He, Zhang, Ren, and Sun}]{he2016deep}
Kaiming He, Xiangyu Zhang, Shaoqing Ren, and Jian Sun. 2016.
\newblock Deep residual learning for image recognition.
\newblock In \emph{Proceedings of the IEEE conference on computer vision and
  pattern recognition}, pages 770--778.

\bibitem[{Hill et~al.(2016)Hill, Cho, Korhonen, and
  Bengio}]{hill_learning_2016}
Felix Hill, KyungHyun Cho, Anna Korhonen, and Yoshua Bengio. 2016.
\newblock \href {https://transacl.org/ojs/index.php/tacl/article/view/711}
  {Learning to {Understand} {Phrases} by {Embedding} the {Dictionary}}.
\newblock \emph{Transactions of the Association for Computational Linguistics},
  4(0):17--30.

\bibitem[{Honnibal and Johnson(2015)}]{honnibal-johnson:2015:EMNLP}
Matthew Honnibal and Mark Johnson. 2015.
\newblock \href {https://aclweb.org/anthology/D/D15/D15-1162} {An improved
  non-monotonic transition system for dependency parsing}.
\newblock In \emph{Proceedings of the 2015 Conference on Empirical Methods in
  Natural Language Processing}, pages 1373--1378, Lisbon, Portugal. Association
  for Computational Linguistics.

\bibitem[{Iyyer et~al.(2015)Iyyer, Manjunatha, Boyd-Graber, and
  Daum{\'e}~III}]{iyyer2015deep}
Mohit Iyyer, Varun Manjunatha, Jordan~L Boyd-Graber, and Hal Daum{\'e}~III.
  2015.
\newblock Deep unordered composition rivals syntactic methods for text
  classification.
\newblock In \emph{ACL (1)}, pages 1681--1691.

\bibitem[{Kingma and Ba(2014)}]{kingma2014adam}
Diederik Kingma and Jimmy Ba. 2014.
\newblock Adam: A method for stochastic optimization.
\newblock \emph{arXiv preprint arXiv:1412.6980}.

\bibitem[{Lampert et~al.(2014)Lampert, Nickisch, and Harmeling}]{lampert}
Christoph~H Lampert, Hannes Nickisch, and Stefan Harmeling. 2014.
\newblock \href {https://doi.org/10.1109/TPAMI.2013.140} {Attribute-based
  classification for zero-shot visual object categorization}.
\newblock \emph{IEEE Transactions on Pattern Analysis and Machine
  Intelligence}, 36(3):453--465.

\bibitem[{Levin(1993)}]{levin1993}
Beth Levin. 1993.
\newblock \emph{English verb classes and alternations : a preliminary
  investigation}.

\bibitem[{Liu et~al.(2011)Liu, Kuipers, and Savarese}]{liu2011recognizing}
Jingen Liu, Benjamin Kuipers, and Silvio Savarese. 2011.
\newblock Recognizing human actions by attributes.
\newblock In \emph{Computer Vision and Pattern Recognition (CVPR), 2011 IEEE
  Conference on}, pages 3337--3344. IEEE.

\bibitem[{Marco and Georgiana(2015)}]{marco2015hubness}
Angeliki~Lazaridou Marco, Baroni and Dinu Georgiana. 2015.
\newblock Hubness and pollution: Delving into cross-space mapping for zero-shot
  learning.
\newblock ACL.

\bibitem[{McRae et~al.(2005)McRae, Cree, Seidenberg, and McNorgan}]{McRae2005}
Ken McRae, George~S Cree, Mark~S Seidenberg, and Chris McNorgan. 2005.
\newblock \href {http://www.ncbi.nlm.nih.gov/pubmed/16629288} {Semantic feature
  production norms for a large set of living and nonliving things}.
\newblock \emph{Behav Res Methods}, 37(4):547--559.

\bibitem[{Mikolov et~al.(2013)Mikolov, Sutskever, Chen, Corrado, and
  Dean}]{mikolov2013distributed}
Tomas Mikolov, Ilya Sutskever, Kai Chen, Greg~S Corrado, and Jeff Dean. 2013.
\newblock Distributed representations of words and phrases and their
  compositionality.
\newblock In \emph{Advances in neural information processing systems}, pages
  3111--3119.

\bibitem[{Miller(1995)}]{miller1995wordnet}
George~A Miller. 1995.
\newblock Wordnet: a lexical database for english.
\newblock \emph{Communications of the ACM}, 38(11):39--41.

\bibitem[{Mukherjee and Hospedales(2016)}]{mukherjee2016gaussian}
Tanmoy Mukherjee and Timothy Hospedales. 2016.
\newblock Gaussian visual-linguistic embedding for zero-shot recognition.
\newblock EMNLP.

\bibitem[{Nilsback and Zisserman(2008)}]{Nilsback08}
Maria-Elena Nilsback and Andrew Zisserman. 2008.
\newblock Automated flower classification over a large number of classes.
\newblock In \emph{Computer Vision, Graphics \& Image Processing, 2008.
  ICVGIP'08. Sixth Indian Conference on}, pages 722--729. IEEE.

\bibitem[{Pedregosa et~al.(2011)Pedregosa, Varoquaux, Gramfort, Michel,
  Thirion, Grisel, Blondel, Prettenhofer, Weiss, Dubourg et~al.}]{scikit-learn}
Fabian Pedregosa, Ga{\"e}l Varoquaux, Alexandre Gramfort, Vincent Michel,
  Bertrand Thirion, Olivier Grisel, Mathieu Blondel, Peter Prettenhofer, Ron
  Weiss, Vincent Dubourg, et~al. 2011.
\newblock Scikit-learn: Machine learning in python.
\newblock \emph{Journal of Machine Learning Research}, 12(Oct):2825--2830.

\bibitem[{Pennington et~al.(2014)Pennington, Socher, and
  Manning}]{pennington2014glove}
Jeffrey Pennington, Richard Socher, and Christopher~D Manning. 2014.
\newblock Glove: Global vectors for word representation.
\newblock In \emph{EMNLP}, volume~14, pages 1532--1543.

\bibitem[{Rohrbach et~al.(2015)Rohrbach, Rohrbach, Tandon, and
  Schiele}]{rohrbach_dataset_2015}
Anna Rohrbach, Marcus Rohrbach, Niket Tandon, and Bernt Schiele. 2015.
\newblock \href {http://arxiv.org/abs/1501.02530} {A {Dataset} for {Movie}
  {Description}}.
\newblock \emph{arXiv:1501.02530 [cs]}.
\newblock ArXiv: 1501.02530.

\bibitem[{Romera-Paredes and Torr(2015)}]{romera2015embarrassingly}
Bernardino Romera-Paredes and Philip~HS Torr. 2015.
\newblock An embarrassingly simple approach to zero-shot learning.
\newblock In \emph{ICML}, pages 2152--2161.

\bibitem[{Rubinstein et~al.(2015)Rubinstein, Levi, Schwartz, and
  Rappoport}]{rubinstein-levy-schwartz-rappoport:2015:acl}
Dana Rubinstein, Effi Levi, Roy Schwartz, and Ari Rappoport. 2015.
\newblock How well do distributional models capture different types of semantic
  knowledge?
\newblock In \emph{Proceedings of the 53nd Annual Meeting of the Association
  for Computational Linguistics (Volume 2: Short Papers)}.

\bibitem[{Siegel and McKeown(2000)}]{siegel2000learning}
Eric~V Siegel and Kathleen~R McKeown. 2000.
\newblock Learning methods to combine linguistic indicators: Improving
  aspectual classification and revealing linguistic insights.
\newblock \emph{Computational Linguistics}, 26(4):595--628.

\bibitem[{Silberer et~al.(2013)Silberer, Ferrari, and
  Lapata}]{silberer-ferrari-lapata:2013:ACL2013}
Carina Silberer, Vittorio Ferrari, and Mirella Lapata. 2013.
\newblock \href {http://www.aclweb.org/anthology/P13-1056} {Models of semantic
  representation with visual attributes}.
\newblock In \emph{Proceedings of the 51st Annual Meeting of the Association
  for Computational Linguistics (Volume 1: Long Papers)}, pages 572--582,
  Sofia, Bulgaria. Association for Computational Linguistics.

\bibitem[{Vendler(1957)}]{vendler_verbs_1957}
Zeno Vendler. 1957.
\newblock \href {http://www.jstor.org/stable/2182371} {Verbs and {Times}}.
\newblock \emph{The Philosophical Review}, 66(2):143--160.

\bibitem[{Vendrov et~al.(2016)Vendrov, Kiros, Fidler, and
  Urtasun}]{VendrovArxiv15}
Ivan Vendrov, Ryan Kiros, Sanja Fidler, and Raquel Urtasun. 2016.
\newblock Order-embeddings of images and language.
\newblock In \emph{ICLR}.

\bibitem[{Welinder et~al.(2010)Welinder, Branson, Mita, Wah, Schroff, Belongie,
  and Perona}]{welinder2010caltech}
Peter Welinder, Steve Branson, Takeshi Mita, Catherine Wah, Florian Schroff,
  Serge Belongie, and Pietro Perona. 2010.
\newblock Caltech-ucsd birds 200.

\bibitem[{Xu et~al.(2017)Xu, Hospedales, and Gong}]{xu2017transductive}
Xun Xu, Timothy Hospedales, and Shaogang Gong. 2017.
\newblock Transductive zero-shot action recognition by word-vector embedding.
\newblock \emph{International Journal of Computer Vision}, pages 1--25.

\bibitem[{Yatskar et~al.(2016)Yatskar, Zettlemoyer, and Farhadi}]{imsitu}
Mark Yatskar, Luke Zettlemoyer, and Ali Farhadi. 2016.
\newblock Situation recognition: Visual semantic role labeling for image
  understanding.
\newblock In \emph{Proceedings of the IEEE Conference on Computer Vision and
  Pattern Recognition}, pages 5534--5542.

\end{thebibliography}
\bibliographystyle{emnlp_natbib}
\appendix
\renewcommand{\thesection}{\Alph{section}}
\section{Supplemental}

\subsection*{Implementation details}

Our CNN and BGRU models are built in PyTorch\footnote{\href{http://pytorch.org/}{pytorch.org}}. All of our one-layer neural network models are built in Scikit-learn \citep{scikit-learn} using the provided LogisticRegression class (using one-versus-rest if appropriate). Our neural models use the Adam optimizer \citep{kingma2014adam}, though we weak the default hyperparameters somewhat.

Recall that our dictionary definition model is a bidirectional GRU with a hidden size of 300, with a vocabulary size of 30,000. After pretraining on the Dictionary Challenge, we freeze the word embeddings and apply a dropout rate of $50\%$ before the final hidden layer. We found that such an aggressive dropout rate was necessary due to the small size of the training set. During pretraining, we used a learning rate of $10^{-4}$, a batch size of $64$, and set the Adam parameter $\epsilon$ to the default $1e^{-8}$. During finetuning, we set $\epsilon=1.0$ and the batch size to $32$. In general, we found that setting too low of an $\epsilon$ during finetuning caused our zero-shot models to update parameters too aggressively during the first couple of updates, leading to poor results.

For our CNN models, we pretrained the ResNet-152 (initialized with Imagenet weights) on the training classes of the imSitu dataset, using a learning rate of $10^{-4}$ and $\epsilon=10^{-8}$. During finetuning, we dropped the learning rate to $10^{-5}$ and set $\epsilon=10^{-1}$. We also froze all parameters except for the final ResNet block, and the linear attribute and embedding weights. We also found L2 regularization quite important in reducing overfitting, and we applied regularization at a weight of $10^{-4}$ to all trainable parameters.

%\subsection*{Text examples}

\begin{table*}[h!]
\centering
\begin{tabular}{l | l | p{25mm} |l |l | l | l | l }
&Model& Definition & Social & Aspect & Energy & Time & Body part \\ \hline
\multirow{4}{*}{\rotatebox[origin = c]{90}{shop}} &GT&  \multirow{4}{*}{\parbox{25mm}{To visit stores in search of merchandise or bargains}} &\bf{likely social}&\bf{accomplish.}&\bf{high}&\bf{hours}&\bf{arms,head}\\
&embed& &likely solitary&activity&medium& minutes&\\
&BGRU& &solitary or social&activity&medium& minutes&\\
&BGRU+& &solitary or social&activity&medium& minutes&\\
\hline
\multirow{4}{*}{\rotatebox[origin = c]{90}{mash}} &GT&  \multirow{4}{*}{\parbox{25mm}{To convert malt or grain into mash}} &\bf{likely solitary}&\bf{activity}&\bf{high}& \bf{seconds}&\bf{arms}\\
&embed& &\bf{likely solitary}&\bf{activity}&medium&\bf{seconds}&\bf{arms}\\
&BGRU& &solitary or social&achievement&medium&\bf{seconds}&\bf{arms}\\
&BGRU+& &\bf{likely solitary}&\bf{activity}&\bf{high}&\bf{seconds}&\bf{arms}\\
\hline
\multirow{4}{*}{\rotatebox[origin = c]{90}{photograph}} &GT&  \multirow{4}{*}{\parbox{25mm}{To take a photograph of}} &\bf{solitary or social}&\bf{achievement}&\bf{low}&\bf{seconds}&\bf{arms,head}\\
&embed& &\bf{solitary or social}&accomplish.&medium& minutes&\bf{arms}\\
&BGRU& &\bf{solitary or social}&\bf{achievement}&medium&\bf{seconds}&\bf{arms}\\
&BGRU+& &\bf{solitary or social}&unclear&\bf{low}&\bf{seconds}&\bf{arms}\\
\hline
\multirow{4}{*}{\rotatebox[origin = c]{90}{spew out}} &GT&  \multirow{4}{*}{\parbox{25mm}{eject or send out in large quantities also metaphorical}} &\bf{solitary or social}&\bf{achievement}&\bf{high}&\bf{seconds}&\bf{head}\\
&embed& &likely solitary&\bf{achievement}&medium&\bf{seconds}&\\
&BGRU& &s\bf{solitary or social}&\bf{achievement}&\bf{high}&\bf{seconds}&arms\\
&BGRU+& &likely solitary&\bf{achievement}&medium&\bf{seconds}&\\
\hline
\multirow{4}{*}{\rotatebox[origin = c]{90}{tear}} &GT&  \multirow{4}{*}{\parbox{25mm}{To pull apart or into pieces by force rend}} &\bf{likely solitary}&\bf{achievement}&\bf{low}&\bf{seconds}&\bf{arms}\\
&embed& &solitary or social&\bf{achievement}&medium&\bf{seconds}&\bf{arms}\\
&BGRU& &solitary or social&\bf{achievement}&high&\bf{seconds}&\bf{arms}\\
&BGRU+& &solitary or social&\bf{achievement}&high&\bf{seconds}&\bf{arms}\\
\hline
\multirow{4}{*}{\rotatebox[origin = c]{90}{squint}} &GT&  \multirow{4}{*}{\parbox{25mm}{To look with the eyes partly closed as in bright sunlight}} &\bf{likely solitary}&\bf{achievement}&\bf{low}&\bf{seconds}&\bf{head}\\
&embed& &\bf{likely solitary}&\bf{achievement}&\bf{low}&\bf{seconds}&\bf{head}\\
&BGRU& &\bf{likely solitary}&\bf{achievement}&\bf{low}&\bf{seconds}&\bf{head}\\
&BGRU+& &\bf{likely solitary}&\bf{achievement}&\bf{low}&\bf{seconds}&\bf{head}\\
\hline
\multirow{4}{*}{\rotatebox[origin = c]{90}{shake}} &GT&  \multirow{4}{*}{\parbox{25mm}{To cause to move to and fro with jerky movements}} &\bf{solitary or social}&\bf{activity}&\bf{medium}&\bf{seconds}&\\
&embed& &likely solitary&achievement&\bf{medium}&\bf{seconds}&arms\\
&BGRU& &likely solitary&\bf{activity}&\bf{medium}&\bf{seconds}&\\
&BGRU+& &likely solitary&\bf{activity}&\bf{medium}&\bf{seconds}&\\
\hline
\multirow{4}{*}{\rotatebox[origin = c]{90}{doze}} &GT&  \multirow{4}{*}{\parbox{25mm}{To sleep lightly and intermittently}} &\bf{likely solitary}&\bf{state}&\bf{none}&\bf{minutes}&\bf{head}\\
&embed& &\bf{likely solitary}&achievement&medium&seconds&\\
&BGRU& &\bf{likely solitary}&achievement&low&seconds&\\
&BGRU+& &\bf{likely solitary}&activity&low&seconds&\\
\hline
\multirow{4}{*}{\rotatebox[origin = c]{90}{writhe}} &GT&  \multirow{4}{*}{\parbox{25mm}{To twist as in pain struggle or embarrassment}} &\bf{solitary or social}&\bf{activity}&\bf{high}&\bf{seconds}&\bf{arms,torso}\\
&embed& &likely solitary&\bf{activity}&medium&\bf{seconds}&\\
&BGRU& &likely solitary&\bf{activity}&medium&\bf{seconds}&\bf{arms}\\
&BGRU+& &likely solitary&\bf{activity}&medium&\bf{seconds}&\\
\hline
\end{tabular}
\caption{Example sentences and predicted attributes. Due to space constraints, we only list a few representative attributes and verbs. GT refers to the ground truth annotations. Bolded predictions are correct.}
\label{text-examples}
\end{table*}

\subsection*{Full list of attributes}
The following is a full list of the attributes. In addition to the attributes presented here, we also crowdsourced attributes for the
emotion content of each verb (e.g., happiness, sadness, anger, and surprise). However, we found these annotations to be skewed towards ``no emotion'', since most verbs do not strongly associate with a specific emotion. Thus, we omit them in our experiments.

{\footnotesize
\begin{enumerate}[(1)]
\item Aspectual Classes: one attribute with 5 values:
\begin{enumerate}[(a)]
\item State
\item Achievement
\item Accomplishment
\item Activity
\item Unclear without context
\end{enumerate}
\item Temporal Duration: one attribute with 5 values:
\begin{enumerate}[(a)]
\item Atemporal
\item On the order of seconds
\item On the order of minutes
\item On the order of hours
\item On the order of days
\end{enumerate}
\item Motion Dynamics: One attribute with 5 values:
\begin{enumerate}[(a)]
\item Unclear without context
\item No motion
\item Low motion
\item Medium motion
\item High motion
\end{enumerate}
\item Social Dynamics: One attribute with 5 values:
\begin{enumerate}[(a)]
\item Solitary
\item Likely solitary
\item Solitary or social
\item Likely social
\item Social
\end{enumerate}
\item Transitivity: Three binary attributes:
\begin{enumerate}[(a)]
\item Intransitive: 1 if the verb can be used intransitively
\item Transitive (person): 1 if the verb can be used in the form ``$<$someone$>$''
\item Transitive (object): 1 if the verb can be used in the form ``$<$verb$>$ something''
\end{enumerate}
\item Effects on Arguments: 12 binary attributes
\begin{enumerate}[(a)]
\item Intransitive 1: 1 if the verb is intransitive and the subject moves somewhere
\item Intransitive 2: 1 if the verb is intransitive and the external world changes
\item Intransitive 3: 1 if the verb is intransitive, and the subject's state changes
\item Intransitive 4: 1 if the verb is intransitive, and nothing changes
\item Transitive (obj) 1: 1 if the verb is transitive for objects and the object moves somewhere
\item Transitive (obj) 2: 1 if the verb is transitive for objects and the external world changes
\item Transitive (obj) 3: 1 if the verb is transitive for objects and the object's state changes
\item Transitive (obj) 4: 1 if the verb is transitive for objects and nothing changes
\item Transitive (person) 1: 1 if the verb is transitive for people and the object is a person that moves somewhere
\item `Transitive (person) 2:  1 if the verb is transitive for people and the external world changes
\item Transitive (person) 3:  1 if the verb is transitive for people and if the object is a person whose state changes
\item Transitive (person) 4: 1 if the verb is transitive for people and nothing changes
\end{enumerate}
\item Body Involements: 5 binary attributes
\begin{enumerate}[(a)]
\item Arms: 1 if arms are used
\item Head: 1 if head is used
\item Legs: 1 if legs are used
\item Torso: 1 if torso is used
\item Other: 1 if another body part is used
\end{enumerate}
\end{enumerate}
}

\end{document}